\documentclass{article}
\PassOptionsToPackage{numbers,sort&compress}{natbib}
\usepackage[preprint]{neurips_2025}

\usepackage[utf8]{inputenc}
\usepackage[T1]{fontenc}
\usepackage{amsmath,amssymb,amsthm}
\usepackage{booktabs}
\usepackage{graphicx}
\usepackage{microtype}
\usepackage{needspace}
\usepackage{xcolor}
\usepackage{xspace}
\usepackage{enumitem}
\usepackage{algorithm}
\usepackage{algpseudocode}
\newcommand{\etal}{\emph{et al.}\xspace}
\definecolor{linkblue}{RGB}{34,77,116}
\usepackage[colorlinks,allcolors=linkblue]{hyperref}

\theoremstyle{plain}
\newtheorem{theorem}{Theorem}
\newtheorem{proposition}{Proposition}
\theoremstyle{definition}
\newtheorem{definition}{Definition}

\newcommand{\pass}[1]{\mathrm{pass@}#1}
\newcommand{\maj}[1]{\mathrm{maj@}#1}
\newcommand{\brier}{\mathrm{Brier}}
\newcommand{\phiq}{\phi_{\mathrm{quad}}}
\newcommand{\E}{\mathbb{E}}
\newcommand{\Acal}{\mathcal{A}}
\newcommand{\one}{\mathbb{1}}

\title{Beyond Sharpening: Proper Scoring Policy Optimization\\for Online LLM Reasoning}

\makeatletter
\hypersetup{pdftitle={\@title}}
\makeatother

\author{%
	Zhengyuan~Fan \\
	School of Computer Science, Northwestern Polytechnical University \\
	\texttt{fanzhengyuan@nwpu.edu.cn}
}
\begin{document}
\maketitle
\begin{abstract}
Online reinforcement learning methods such as GRPO have become the main way to improve the reasoning of large language models, yet it is still debated whether their gains come mainly from sharpening, that is, from making the model's existing answer distribution more concentrated. The answer decides how these methods can be improved. We argue that GRPO does sharpen the output distribution, and that the sharpening comes from its training objective, pass@1 (the probability that a single sample is correct), itself. To show this, we study sharpening in reverse, through the single-sample error rate, which we decompose exactly into four layers: errors removed by lowering the sampling temperature, errors removed by majority voting, errors where the correct answer is sampled but is not the most frequent answer, and errors where the correct answer is never sampled. Measurements show that GRPO reduces errors almost only in the first two layers, while the third layer grows and the fourth is unchanged: GRPO sharpens the most frequent answer even when it is wrong, and lowers the chance of sampling correct answers that are in the minority. To counter this effect, we view the distribution of sampled answers as a probabilistic prediction of the correct answer. From this viewpoint the objective splits exactly as $\pass{1}=\tfrac12\brier+\tfrac12\sum_a q(a)^2$. The first term is the Brier score, which measures the prediction error and is maximized exactly at the true answer distribution; the second is half the probability that two independent samples give the same answer, and it ignores the label. This shows that GRPO and its variants sharpen the problems whose most frequent answer is already correct, at the expense of correct answers in the minority. We therefore propose Proper Scoring Policy Optimization (PSPO): for most of training it removes the sharpening term and adjusts answer probabilities with the Brier score alone, and near the end it adds the term back under label guidance, which keeps the model from becoming overconfident and from drifting far from its original distribution. More importantly, the decomposition generalizes: every objective that depends only on the probability of the correct answer, including GRPO, pass@$k$ and MaxRL with a finite group size, splits uniquely into a proper scoring rule plus a label-free term, which gives a common basis for analyzing and designing such objectives. In a simulation where discriminative features are learned more slowly than sharpening, PSPO lowers the fraction of third-layer problems that fall into the fourth layer from 21\% to 0\% compared with GRPO and raises pass@16 from 0.866 to 0.978; two-stage PSPO is slightly more accurate than GRPO at $T=0.6$ while keeping pass@16 9.2 points higher. Problems in the fourth layer stay out of reach for any proper score and still require distillation or external search.
\end{abstract}

\section{Introduction}
\label{sec:intro}

Online reinforcement learning with verifiable rewards (RLVR) has become the main way to improve the reasoning of large language models, with GRPO and its variants as the representative algorithms~\cite{shao2024deepseekmath,guo2025deepseekr1,liu2025drgrpo}. The central debate around it is whether these methods teach the model new reasoning abilities, or only make the answer distribution the model already has more concentrated, which is called \emph{sharpening}~\cite{yue2025rl,zhao2025echo,wu2026leash}. The answer decides where improvements should come from. Compared with offline methods that rely on a teacher or external search, online methods use only the model's own samples and are cheap, but in practice they mainly improve problems the model can already solve, and long training can even lower pass@$k$~\cite{yue2025rl,song2025obe}. We study the simplest online setting: a 0/1 answer verifier, no teacher, and no extra samples.

\begin{figure}[t]
	\centering
	\includegraphics[width=\linewidth]{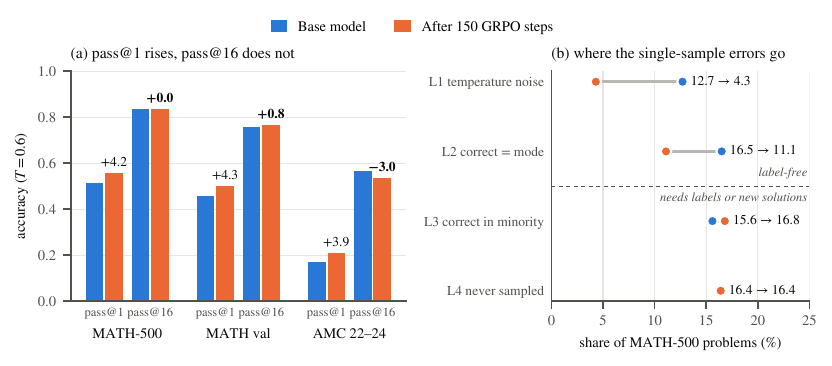}
	\caption{What GRPO changes. \texttt{Qwen2.5-1.5B-Instruct} before and after 150 GRPO steps, 16 samples per problem. (a) At $T=0.6$, pass@1 rises by about 4 points on every set, while pass@16 does not rise and falls by 3.0 points on the harder AMC 22--24; numbers are changes in points. (b) The four-layer split of the single-sample error (Equation~\ref{eq:layers}, $T_0=1.0$, $T_1=0.6$, $k=16$) on MATH-500: the gains are entirely in the two label-free layers, while the layer where the correct answer is in the minority grows.}
	\label{fig:insight}
\end{figure}

We argue that GRPO does sharpen the output distribution, and that the sharpening comes mainly from its training objective, pass@1, the probability that a single sample is correct. Sharpening itself is hard to measure directly, so we approach it from the other side and ask which single-sample errors training removes. To this end we combine pass@1, majority-vote accuracy maj@$k$ and pass@$k$, and decompose the single-sample error rate exactly into four layers (Section~\ref{sec:diagnosis}): errors removed by lowering the sampling temperature (L1); errors where the correct answer is already the most frequent answer, which majority voting removes (L2); errors where the correct answer can be sampled but is not the most frequent one, i.e., it is in the minority (L3); and errors where the correct answer is never sampled in $k$ tries (L4). The first two layers need no label, the third needs a label, and the fourth needs information outside the sampling distribution. On \texttt{Qwen2.5-1.5B-Instruct}, 150 steps of GRPO lower the L1 and L2 errors on MATH-500 from 12.7\% and 16.5\% to 4.3\% and 11.1\%, while L3 grows from 15.6\% to 16.8\% and L4 is unchanged (Figure~\ref{fig:insight}). In other words, GRPO sharpens the most frequent answer even when it is wrong, and lowers the chance of sampling correct answers in the minority; accordingly pass@1 rises while pass@16 stays flat or drops. Our diagnostics on five to eight models further show that refining the reward at the level of tokens or reasoning steps does not change this (Section~\ref{sec:process}).

To counter this effect, we first need to understand why pass@1 sharpens in this way. We view the probability distribution of sampled answers as a probabilistic prediction of the correct answer: training should bring this prediction close to the true distribution of correct answers (Section~\ref{sec:bridge}). The standard measure of the error of such a prediction is the Brier score~\cite{brier1950verification}, whose optimum is exactly the true distribution; scores with this property are called strictly proper scoring rules~\cite{gneiting2007strictly}. From this viewpoint, the training objective splits exactly into two terms (Section~\ref{sec:identity}):
\begin{equation*}
	\pass{1} = \tfrac12\,\brier + \tfrac12\textstyle\sum_a q(a)^2 .
\end{equation*}
The first term is half the Brier score. The second is half the probability that two independent samples give the same answer; it ignores the label, and we call it the \emph{sharpening term}: it grows whenever probability concentrates on one answer, whether that answer is right or wrong.

This split shows which part GRPO and its variants sharpen. For most problems the correct answer is already the most frequent answer. On these problems the sharpening term agrees with the label, and the more dominant the correct answer, the larger the share of the update that comes from the sharpening term; this is where most of GRPO's gain comes from. On L3 problems, by contrast, the sharpening term rewards the most frequent wrong answer. Because all problems share parameters, the aggregate update is almost purely in the sharpening direction, and L3 problems drift into L4. Grouping problems into classes makes the mechanism explicit (Section~\ref{sec:class}): for a class of problems the model cannot yet tell apart, the optimum of pass@1 puts all probability on the most frequent answer of the class, whereas without the sharpening term the optimum is exactly the true distribution of correct answers in the class. Only this square term achieves that; a cubic term or a different coefficient does not.

We therefore propose \textbf{Proper Scoring Policy Optimization (PSPO)} (Section~\ref{sec:method}). For most of training it removes the sharpening term and adjusts answer probabilities with the Brier score alone; near the end it adds the sharpening term back under label guidance, raising its weight from 0 to 1 to improve single-sample accuracy. Because the optimum of the Brier score is the true distribution rather than the most frequent answer, during the main stage the model neither becomes overconfident nor drifts far from its original distribution by collapsing onto a single answer. Subtracting the group mean as GRPO does would reintroduce a sharpening term with weight $1/(G-1)$; we give a leave-one-out correction that keeps PSPO unbiased for any group size $G\ge 3$, with the same sampling as GRPO and no teacher. The reward of PSPO resembles the penalty on repeated answers of Song \etal~\cite{song2025obe}; the differences are that the coefficient of the penalty is fixed by proper scoring (exactly 1 relative to the 0/1 reward), that the advantage is not divided by the group standard deviation, and that the sharpening term is added back under labels at the end (Section~\ref{sec:related}).

More importantly, the split generalizes (Section~\ref{sec:general}): every objective that depends only on the probability of the correct answer, including GRPO, pass@$k$ and MaxRL with a finite group size~\cite{tajwar2026maxrl}, splits uniquely into a proper scoring rule plus a label-free term. Only the logarithmic score has no label-free term, but it cannot be estimated without bias from finitely many samples; label-free methods such as TTRL~\cite{zuo2025ttrl} contain only the label-free term. Labeled and label-free online methods can thus be compared in one framework, and a new objective can be checked simply by asking how its label-free term varies with the probability of an answer. Section~\ref{sec:exp} verifies these results in simulations with shared parameters. A proper score can correct L3, but it cannot reach L4, which still requires offline distillation or external search.

Our contributions are:
\begin{itemize}[leftmargin=*,itemsep=1pt,topsep=2pt]
	\item an exact four-layer decomposition of the single-sample error, and diagnostics of GRPO and of process-level interventions across several models (Section~\ref{sec:diagnosis});
	\item the identity $\pass{1}=\tfrac12\brier+\tfrac12\sum_a q(a)^2$, obtained by viewing the answer distribution as a probabilistic prediction. It shows that GRPO sharpens the part where the correct answer is already in the majority, explains why L3 problems fall into L4, and is the unique such split under the requirement that the remainder be strictly proper (Section~\ref{sec:theory});
	\item PSPO, with a leave-one-out advantage that is unbiased for any $G\ge3$ (Section~\ref{sec:method});
	\item a generalization that splits GRPO and its variants uniquely into a proper score and a label-free part, and a proof that MaxRL with a finite group size suppresses low-frequency answers (Section~\ref{sec:general});
	\item verification in three simulated environments with shared parameters (Section~\ref{sec:exp}).
\end{itemize}

\section{Diagnosis: which errors does online RL remove?}
\label{sec:diagnosis}

\subsection{A four-layer decomposition of the single-sample error}
\label{sec:layers}

Given a problem $x$, let $\pi_{\theta,T}(y\mid x)$ be the distribution of responses $y$ at sampling temperature $T$. The verifier maps each response to a final answer $a(y)\in\Acal$, with mathematically equivalent answers treated as one; a response without a \verb|\boxed{}| answer or cut at the length limit is mapped to an empty answer $\varnothing\in\Acal$. This gives the model's \emph{answer distribution} and the probability of a correct answer,
\begin{equation}
	q_{\theta,T}(a\mid x)\triangleq\sum_y \pi_{\theta,T}(y\mid x)\,\one[a(y)=a],\qquad p_T(x)\triangleq q_{\theta,T}(a^*\mid x),
\end{equation}
where $a^*$ is the reference answer, so that $\pass{1}_T=\E_x[p_T(x)]$. For $T=1$ we write $q_\theta$ and $p$. Let $\maj{k}_T$ be the accuracy of a majority vote over $k$ independent samples (ties broken at random) and $\pass{k}_T=\E_x[1-(1-p_T(x))^k]$.

For any $T_0\ge T_1>0$ and $k\ge1$, the single-sample error rate is exactly the sum of four terms:
\begin{equation}
	\begin{aligned}
	1-\pass{1}_{T_0}
	={}&\underbrace{\big(\pass{1}_{T_1}-\pass{1}_{T_0}\big)}_{\Delta_1}
	+\underbrace{\big(\maj{k}_{T_1}-\pass{1}_{T_1}\big)}_{\Delta_2}\\
	&+\underbrace{\big(\pass{k}_{T_1}-\maj{k}_{T_1}\big)}_{\Delta_3}
	+\underbrace{\big(1-\pass{k}_{T_1}\big)}_{\Delta_4}.
	\end{aligned}
	\label{eq:layers}
\end{equation}
The four terms are the gains of four inference-time operations: $\Delta_1$ is the gain from lowering the temperature from $T_0$ to $T_1$; $\Delta_2$ is the gain from a majority vote over $k$ samples; $\Delta_3$ is what could still be gained if the reference answer were used to pick a correct sample among the $k$; and $\Delta_4$ is the part where all $k$ samples are wrong, which only information from outside the model can fix.

At the level of single problems, the answer distribution $q_{\theta,T_1}(\cdot\mid x)$ puts each problem (ignoring exact ties) into one of four classes: \emph{solved}, if $a^*$ is the mode and $q_{\theta,T_1}(a^*\mid x)\approx1$; \emph{L2}, if $a^*$ is the mode but clearly below 1; \emph{L3}, if $0<q_{\theta,T_1}(a^*\mid x)<\max_{a\ne a^*}q_{\theta,T_1}(a\mid x)$, so that the correct answer is in the minority; and \emph{L4}, if $q_{\theta,T_1}(a^*\mid x)\approx0$. The two views agree: for large $k$ and no ties, $\Delta_3\approx\Pr(x\in\text{L3})$, $\Delta_4\approx\Pr(x\in\text{L4})$, and $\Delta_2$ is the net effect of majority voting, its gain on L2 minus its loss on L3. Section~\ref{sec:exp} tracks how individual problems move between classes during training, e.g.\ from L3 to L4 (written L3$\to$L4). Table~\ref{tab:layers} summarizes the layers.

\begin{table}[t]
	\centering
	\caption{The four layers of the single-sample error. We use $T_0=1.0$, $T_1=0.6$, $k=16$ throughout.}
	\label{tab:layers}
	\small
	\begin{tabular}{p{0.27\linewidth}p{0.32\linewidth}p{0.31\linewidth}}
		\toprule
		Layer & Condition on $q=q_{\theta,T_1}(\cdot\mid x)$ & What removes it \\
		\midrule
		L1: temperature noise & token-level sampling leaves high-probability paths & lower temperature; no label \\
		L2: correct answer is the mode & $a^*=\arg\max_a q(a)$, $q(a^*)<1$ & concentrate the distribution; no label \\
		L3: correct answer in the minority & $0<q(a^*)<\max_{a\ne a^*}q(a)$ & demote the common wrong answer; needs the label \\
		L4: never sampled & $q(a^*)\approx0$ & new solutions (distillation, external search) \\
		\bottomrule
	\end{tabular}
\end{table}

We use $T_0=1.0$ because the policy gradient of online RL acts on the $T=1$ distribution; RL can mimic a lower temperature simply by scaling the logits, and keeping $T_0=1.0$ separates this gain. We do not take $T_1$ to be greedy decoding: greedy decoding admits no multi-sample metrics, and reasoning models often loop or hit the length limit under greedy decoding (Appendix~\ref{app:diag}). L1 and L2 are both label-free errors of an insufficiently concentrated distribution, and the theory in Sections~\ref{sec:theory}--\ref{sec:general} treats them together.

\subsection{Which errors does GRPO actually remove?}
\label{sec:grpo}

We ran online RL on 8,086 deduplicated MATH Level 3--5 training problems with three kinds of models and examined the result with the decomposition.

\textbf{Instruction model.} \texttt{Qwen2.5-1.5B-Instruct}, 150 GRPO steps, 32 problems and 8 samples per step, evaluated with 16 samples per problem. On MATH-500, pass@1 at $T=1.0$ rises from 0.388 to 0.514 and at $T=0.6$ from 0.515 to 0.557. By Equation~\eqref{eq:layers} the whole net gain is in the first two layers: L1 falls from 12.7\% to 4.3\% and L2 from 16.5\% to 11.1\%, while maj@16 slightly drops from 0.680 to 0.668, L3 grows from 15.6\% to 16.8\%, and L4 (16.4\%) and pass@16 (0.836) do not change. On five test sets (MATH-500, AMC 22--24, a 500-problem MATH validation set, GSM8K and MMLU-Pro), greedy accuracy improves by 0.6--4.2 points, but the trained model's pass@1 at $T=0.6$ stays within $[-0.8,+1.4]$ points of the base model's greedy accuracy, so most of the gain amounts to a more concentrated distribution. On the harder AMC 22--24, pass@16 falls from 0.567 to 0.537, and on GSM8K pass@4 falls from 0.894 to 0.868.

\textbf{Math base model and long chain-of-thought model.} On \texttt{Qwen2.5-Math-1.5B}, GRPO raises greedy accuracy on MATH-500 from 0.094 to 0.634 in 25 steps, and a method that updates only about 40 forking tokens per step reaches 0.636 in the same steps, mostly by changing the first tokens so that the model stops repeating the problem; the gain comes from fixing repetition loops. On \texttt{DeepSeek-R1-Distill-Qwen-1.5B} (60 steps, 2 seeds), GRPO gains 1.0--5.5 points under the raw evaluation, but once the model is forced to answer at the length limit (budget forcing, no training), the gap to the base model shrinks to $-0.9$ to $+1.5$ points; the gain comes from learning to answer before the length limit (Appendix~\ref{app:diag}).

Most problems already have the correct answer as their mode. Table~\ref{tab:baselines} lists the four layers of five models on a 300-problem MATH analysis set (m300). Across these models, among the problems with both correct and wrong samples, L3 accounts for only 4.2\%--21.7\% (3.2\%--25.3\% on the validation set), and the rest is mostly L2. This explains why average pass@1 rises while pass@$k$ stalls or falls. L2 problems quickly approach full accuracy as the distribution concentrates, but their pass@16 was already close to 1; L3 problems have a small probability of the correct answer, which the overall sharpening moves to the most frequent wrong answer, so even many samples find the correct answer less and less often.

\begin{table}[t]
	\centering
	\caption{Four-layer baselines of five models on m300 (16 samples per problem) and the change under GRPO on MATH-500. The first four columns are accuracies, the last four are layer sizes in points.}
	\label{tab:baselines}
	\small
	\setlength{\tabcolsep}{4.2pt}
	\resizebox{\linewidth}{!}{%
	\begin{tabular}{lcccccccc}
		\toprule
		Model & pass@1 ($T{=}1$) & pass@1 ($T{=}0.6$) & maj@16 & pass@16 & L1 & L2 & L3 & L4 \\
		\midrule
		\multicolumn{9}{l}{\emph{m300, raw evaluation}} \\
		Qwen2.5-1.5B-Instruct & 0.303 & 0.436 & 0.604 & 0.773 & 13.3 & 16.8 & 16.9 & 22.7 \\
		Qwen2.5-3B-Instruct & 0.523 & 0.590 & 0.702 & 0.847 & 6.6 & 11.2 & 14.5 & 15.3 \\
		Llama-3.2-3B-Instruct & 0.241 & 0.392 & 0.712 & 0.820 & 15.1 & 32.0 & 10.8 & 18.0 \\
		Qwen3-4B (non-thinking) & 0.773 & 0.781 & 0.869 & 0.923 & 0.8 & 8.8 & 5.5 & 7.7 \\
		R1-Distill-Qwen-1.5B & 0.714 & 0.733 & 0.893 & 0.927 & 1.9 & 16.1 & 3.3 & 7.3 \\
		\quad forced answer, $T_0{=}T_1{=}0.6$ & -- & 0.802 & 0.883 & 0.943 & -- & 8.1 & 6.0 & 5.7 \\
		\midrule
		\multicolumn{9}{l}{\emph{MATH-500, before and after 150 GRPO steps}} \\
		Qwen2.5-1.5B-Instruct (base) & 0.388 & 0.515 & 0.680 & 0.836 & 12.7 & 16.5 & 15.6 & 16.4 \\
		Qwen2.5-1.5B-Instruct (GRPO) & 0.514 & 0.557 & 0.668 & 0.836 & \textbf{4.3} & \textbf{11.1} & \textbf{16.8} & 16.4 \\
		\bottomrule
	\end{tabular}}
\end{table}

\subsection{Why not intervene at the level of tokens or steps?}
\label{sec:process}

Faced with L3 not improving, a common idea is that answer-level rewards are too sparse and credit should be assigned more finely to key tokens or key steps. We tested this direction on five models (\texttt{Qwen2.5-1.5B/3B-Instruct}, \texttt{Llama-3.2-3B-Instruct}, \texttt{Qwen3-4B} in non-thinking mode and \texttt{DeepSeek-R1-Distill-Qwen-1.5B}) and on sampling data from eight models (details in Appendix~\ref{app:process}). (i) Failure causes differ widely across models, so there is no common process-level rule. (ii) Correct and wrong continuations separate at the level of segments rather than single tokens: the variance of continuation success across segments is 1.51--3.93 times that across tokens, and a single token that turns a greedy continuation from wrong to right raises the success rate of $T=1$ continuations by only $-0.01$ to $+0.04$; a method that updates only forking tokens reaches 0.590 on MATH-500 versus 0.641 for GRPO. (iii) There is no reliable signal before the fact: token entropy, cheap text signals and linear probes on hidden states identify key steps with AUROC 0.51--0.66, below the 0.75 needed in practice. (iv) Even step-wise search guided by the reference answer loses to majority voting at equal compute: at a cost comparable to 16 independent samples it is $-9.6$ to $+2.8$ points from maj@16, and with 2.8--8.3 times the compute all five models remain below pass@16 at 1$\times$ compute. Fine-grained intervention during generation thus has no reliable signal and uses compute less efficiently than aggregating at the answer level.

\subsection{Summary: what should the answer distribution learn?}
\label{sec:bridge}

The diagnostics show two things. GRPO gains by making the answer distribution more concentrated, at the cost of pushing out correct answers in L3; and refining rewards at the level of tokens or steps does not fix this, because the problem lies in the training objective on the answer distribution. What, then, should the answer distribution be trained to be?

Consider a class of problems that the model cannot yet tell apart. It can only give them a common answer distribution $q$, while the correct answers of the class follow some true proportions $b$. For example, 70\% of the problems are solved by the usual method, giving answer $A$, while the other 30\% have a subtle condition and the answer is $B$. The best choice is to report $q=b$ honestly: the 30\% of problems with answer $B$ can still be sampled correctly, so pass@$k$ loses nothing and the model still has correct samples to learn from once it learns to separate the two kinds of problems; and the mode is still $A$, so majority voting (or a lower temperature) gives the same single-answer accuracy as betting on $A$. Conversely, an objective that pushes $q$ to bet only on $A$ moves the 30\% from L3 to L4, after which their correct answers are hard to ever sample again. We therefore want an objective whose expected score, when correct answers follow $b$, is maximized at $q=b$. Such scores are called strictly proper scoring rules~\cite{gneiting2007strictly}. Section~\ref{sec:theory} views the answer distribution as a probabilistic prediction and checks pass@1 against this criterion.

\section{Theory: the sharpening term in pass@1}
\label{sec:theory}

\subsection{The answer distribution as a probabilistic prediction}
\label{sec:prediction}

We collect all outputs of the model on a problem into the answer distribution $q_\theta(\cdot\mid x)\in\Delta(\Acal)$ and view it as a probabilistic prediction of the unknown correct answer $a^*$ (a probabilistic forecast in the terminology of~\cite{gneiting2007strictly}). By Section~\ref{sec:bridge}, training should bring this prediction close to the true distribution of correct answers. The standard measure of the error of a probabilistic prediction is the Brier loss~\cite{brier1950verification}, the squared error between the prediction and the outcome, $\|e_{a^*}-q\|_2^2=\sum_a(q(a)-\one[a=a^*])^2$, where $e_{a^*}$ is the unit vector of the correct answer. To match the direction of pass@1 we use the Brier score, which is larger for better predictions and equals 1 for a perfect one:
\begin{equation*}
	\brier(q,a^*)\triangleq1-\|e_{a^*}-q\|_2^2=2q(a^*)-\textstyle\sum_a q(a)^2 .
\end{equation*}

\begin{definition}[Strictly proper scoring rule~\cite{gneiting2007strictly}]
A scoring rule $S:\Delta(\Acal)\times\Acal\to\mathbb{R}\cup\{-\infty\}$ is strictly proper if, for every $b\in\Delta(\Acal)$, the expected score $\E_{a^*\sim b}[S(q,a^*)]$ is maximized only at $q=b$.
\end{definition}
That is, if correct answers follow $b$, reporting $q=b$ honestly gives the highest expected score. Both the Brier score and the logarithmic score $\log q(a^*)$ are strictly proper.

\subsection{An identity for pass@1}
\label{sec:identity}

From this viewpoint pass@1 has a problem, which a rearrangement of the definition of the Brier score makes visible.

\begin{theorem}[Decomposition of pass@1]
\label{thm:identity}
For any answer distribution $q_\theta(\cdot\mid x)$ and correct answer $a^*$,
\begin{equation}
	\pass{1}(x)=q_\theta(a^*\mid x)
	=\underbrace{\tfrac12\brier\big(q_\theta(\cdot\mid x),a^*\big)}_{\text{strictly proper, }S_{\brier}}
	+\underbrace{\tfrac12\textstyle\sum_{a\in\Acal}q_\theta(a\mid x)^2}_{\text{sharpening term }\phiq=\frac12\Pr(a_1=a_2)},
	\label{eq:identity}
\end{equation}
where $a_1,a_2$ are the answers of two independent samples. Under the requirement that the remainder be strictly proper and the subtracted part be independent of the label, this split is unique up to a constant. Correspondingly, the policy gradient is
\begin{equation}
	\nabla_\theta q_\theta(a^*\mid x)=\E_{y\sim\pi_\theta(\cdot\mid x)}\Big[\big(\underbrace{r(y)-q_\theta(a(y)\mid x)}_{\text{depends on the label}}+\underbrace{q_\theta(a(y)\mid x)}_{\text{label-free}}\big)\nabla_\theta\log\pi_\theta(y\mid x)\Big].
	\label{eq:identity_grad}
\end{equation}
\end{theorem}

The proof is in Appendix~\ref{app:identity}, and Section~\ref{sec:class} explains the uniqueness. pass@1 is thus the sum of the Brier score and a sharpening term. The sharpening term is half the probability that two independent samples agree: it grows whenever probability concentrates on one answer, regardless of whether that answer is correct. At the level of single samples, every 0/1 reward $r(y)$ equals $[r(y)-q(a(y))]+q(a(y))$. The second part is the share of the sample's answer in the current distribution and never looks at the reference answer; it is a smoothed version of the objective of label-free methods such as TTRL~\cite{zuo2025ttrl}, which use the majority answer as a pseudo-label. Optimizing pass@1 therefore always adds a label-free sharpening reward with weight 1 on top of a proper score.

\textbf{The sharpening term acts differently on different layers.} The derivative of the square term with respect to an answer's probability is that probability itself, so the more concentrated the answers, the stronger its push.
\begin{itemize}[leftmargin=*,itemsep=1pt,topsep=2pt]
	\item \emph{L2.} The mode is the correct answer. The sharpening term gives the correct answer the largest reward and agrees with the label, and the more dominant the correct answer, the larger the share of the update it provides. With the correct answer at 0.70 and six wrong answers at 0.05 each, the advantage $+0.30$ that Dr.GRPO gives the correct answer consists of $+0.195$ from the sharpening term and only $+0.105$ from the Brier score (Section~\ref{sec:objective}); with a single wrong answer the share of the sharpening term is $2p-1$, more than half once $p>3/4$.
	\item \emph{L3.} The mode is wrong. With the correct answer at 0.15 and the most frequent wrong answer at 0.50, the sharpening term gives the wrong answer $+0.50$ and the correct one only $+0.15$, against the label, weakening or cancelling the label-dependent part.
	\item \emph{L4.} When the correct answer is never sampled, $r\equiv0$ and the label-dependent part $-q(a)$ cancels the sharpening term $+q(a)$ exactly, so pass@1 has no gradient. Without the sharpening term, the Brier score still pushes down answers that keep recurring (Section~\ref{sec:objective}).
	\item \emph{Parameter sharing amplifies the effect.} Under a natural-gradient approximation, the change of a problem's answer entropy is $\Delta H(x)\approx-\eta\,\mathrm{Cov}_{a\sim q}(\log q(a),A(a))$, the same form as the token-level rule of Cui \etal~\cite{cui2025entropy}. The covariance is positive on L2 problems, so their entropy falls; on L3 problems it is negative when $p<e^{-H(q)}$, so their entropy tends to rise. Yet among problems with both correct and wrong samples, L3 is only 4.2\%--21.7\% (m300; 3.2\%--25.3\% on the validation set): for most problems the correct answer is already the mode. Summed over problems through shared parameters, the update is almost purely sharpening; acting on the L3 problems of the test set, it further lowers the probability of correct answers in the minority, so the model can hardly learn these minority correct paths and the problems fall into L4.
\end{itemize}

\subsection{Grouping problems into classes: why the square term}
\label{sec:class}

To see these results more directly, we partition the data into classes $Z$ (for instance by difficulty or by problem type) and write the RLVR objective as
\begin{equation*}
	J(\theta)=\E_Z\Big[\E_{x\mid Z}\big[\E_{y\sim\pi_\theta(\cdot\mid x)}[r(y)]\big]\Big]=\E_Z\Big[\E_{x\mid Z}\big[q_\theta(a^*(x)\mid x)\big]\Big].
\end{equation*}
This holds for any partition. The most revealing one groups problems the model cannot yet tell apart (Section~\ref{sec:bridge}): the model can only output a common distribution $q(\cdot\mid Z)$ for the class, while the correct answers of the class follow $b(\cdot\mid Z)$; we drop $Z$ when no confusion arises. Within the class, pass@1 equals $\sum_a b(a)q(a)$, which is linear in $q$. Averaging Equation~\eqref{eq:identity} over the class gives
\begin{equation}
	\sum_a b(a)\,q(a)=\underbrace{\tfrac12\textstyle\sum_a b(a)^2}_{\text{fixed by the class}}-\underbrace{\tfrac12\|q-b\|_2^2}_{\text{prediction error}}+\underbrace{\tfrac12\textstyle\sum_a q(a)^2}_{\text{sharpening term}}.
	\label{eq:class}
\end{equation}
The first term does not depend on how the model allocates probability and can only be raised by learning new features that separate the class; the second is zero if and only if $q=b$; the third is the sharpening term. Without the sharpening term the optimum is exactly $q=b$: in the example of Section~\ref{sec:bridge}, the 30\% of problems with answer $B$ can still be sampled correctly, and the mode is still $A$. With it, i.e.\ optimizing pass@1 itself, the optimum is $q=\delta_{a_0}$ whenever the mode $a_0=\arg\max_a b(a)$ is unique: all probability goes to the mode and every other answer, including correct answers in the minority, gets probability 0. This is the limiting case of L3 falling into L4. On a single problem ($b=e_{a^*}$) the Brier score and the sharpening term have the same optimum, betting on the correct answer; the conflict appears only among problems the model cannot tell apart.

\textbf{Why the subtracted term must be the square.} Algebraically, any term can be subtracted from pass@1 and added back. We require, however, that the subtracted part $X(q)$ ignore the label and that the remainder $q(a^*)-X(q)$ be strictly proper. Within a class the remainder has expectation $\sum_a b(a)q(a)-X(q)$. For it to be maximized at $q=b$ under $\sum_a q(a)=1$, the first-order condition requires $b(a)-\partial X/\partial q(a)$ to be the same for all answers $a$. The derivative of the first term with respect to $q(a)$ is exactly $b(a)$; since this must hold for every $b$, we need $\partial X/\partial q(a)=q(a)+\text{const}$, i.e.\ $X=\tfrac12\sum_a q(a)^2$ up to a constant. Neither a cubic term nor a different coefficient works. For $b=(0.625,0.125,0.25)$, the remainder's best report is $(1,0,0)$ for a constant $X$ (pass@1 itself); about $(0.69,0,0.31)$ for $X=\tfrac13\sum_a q(a)^3$, which drives the answer with probability 0.125 to zero; $(0.875,0,0.125)$ for $X=\tfrac14\sum_aq(a)^2$, more concentrated than the truth; about $(0.48,0.23,0.29)$ for $X=\sum_aq(a)^2$, flatter than the truth; and exactly $(0.625,0.125,0.25)$ for $X=\tfrac12\sum_aq(a)^2$.

\textbf{General proper scores.} Equation~\eqref{eq:class} is the Brier case of a general fact. By Savage's representation~\cite{savage1971elicitation,gneiting2007strictly}, every strictly proper score is generated by a strictly convex function $\phi$ on the simplex,
\begin{equation*}
	S_\phi(q,a^*)=\phi(q)+\big\langle\nabla\phi(q),\,e_{a^*}-q\big\rangle;
\end{equation*}
$\phi(q)=\sum_a q(a)\log q(a)$ gives the logarithmic score and $\phi(q)=\tfrac12\sum_aq(a)^2$ gives half the Brier score. What a proper score can and cannot do follows from the decomposition below (proof in Appendix~\ref{app:bregman}).

\begin{proposition}[Bregman decomposition]
\label{prop:bregman}
Let $D_\phi(u\|v)\triangleq\phi(u)-\phi(v)-\langle\nabla\phi(v),u-v\rangle\ge0$ be the Bregman divergence of $\phi$. For $a^*\sim b$,
\begin{equation*}
	\E_{a^*\sim b}\big[\phi(e_{a^*})-S_\phi(q,a^*)\big]=\underbrace{\E_{a^*\sim b}[\phi(e_{a^*})]-\phi(b)}_{\text{generalized conditional entropy }H_\phi(A^*\mid Z)}+\underbrace{D_\phi(b\,\|\,q)}_{\text{calibration error, zero iff }q=b}.
\end{equation*}
\end{proposition}
For $\phi(q)=\sum_aq(a)\log q(a)$ this is the cross-entropy split $H(A^*\mid Z)+\mathrm{KL}(b\|q)$; for $\phi(q)=\tfrac12\sum_aq(a)^2$ it is $\tfrac12(1-\sum_ab(a)^2)+\tfrac12\|b-q\|_2^2$, in agreement with Equation~\eqref{eq:class}. A proper score can drive the calibration error to zero; the first term is set by the representation and can only be lowered by learning better features, for which a proper score keeps low-probability answers within reach.

\section{Method: PSPO}
\label{sec:method}

\subsection{Objective: remove the sharpening term, add it back at the end}
\label{sec:objective}

By Theorem~\ref{thm:identity}, pass@1 is the Brier score plus a sharpening term with weight fixed at 1. PSPO makes this weight a controllable $\kappa$:
\begin{equation}
	J_{\mathrm{PSPO}}(\kappa)=\E_x\big[S_{\brier}(q_\theta(\cdot\mid x),a^*)\big]+\kappa\,\E_x\big[\phiq(q_\theta(\cdot\mid x))\big],\qquad\kappa\in[0,1],
	\label{eq:pspo}
\end{equation}
with per-sample reward $r(y)-(1-\kappa)\,q(a(y))$. For $\kappa=0$ only the Brier score remains and the optimum is the true distribution; $\kappa=1$ recovers pass@1 (Dr.GRPO). Training uses $\kappa=0$ for its main stage and raises $\kappa$ to 1 at the end (Section~\ref{sec:twostage}). Because the optimum of the Brier score is the true distribution rather than the mode, the first stage does not push probability onto a single answer: the model does not become overconfident, and its answer distribution does not drift far from the original one by collapsing. In the three simulated environments of Section~\ref{sec:exp}, the answer entropy at the end of the first stage is 1.06, 0.42 and 0.57, against 0.01, 0.17 and 0.20 for GRPO (Table~\ref{tab:toy}).

Among proper scores we choose the Brier score for four reasons.
\begin{enumerate}[leftmargin=*,itemsep=1pt,topsep=2pt]
	\item \emph{It directly demotes the most common wrong answer.} The Brier reward penalizes each wrong answer by its share in the group, so the more common a wrong answer, the larger its penalty, whereas the logarithmic score gives every wrong answer the same $-1$. This matches the statistics of Section~\ref{sec:maxrl}: in 41\%--94\% of the problems with wrong samples, the wrong samples concentrate on one wrong answer.
	\item \emph{All-wrong groups still have a gradient, which also corrects format failures.} When all samples of a group are wrong ($r_i\equiv0$), objectives that depend only on correctness (GRPO, Dr.GRPO, MaxRL) give zero advantage, while the Brier reward keeps the term $-q(a(y))$, pushing down recurring wrong answers and leaving probability to rare solutions. Since responses without a \verb|\boxed{}| answer or cut at the length limit all map to the empty answer $\varnothing$, recurring format failures form one frequent wrong answer and are penalized, which pushes models such as Llama (38\% of responses without an answer box) or R1 (20\%--35\% at the length limit) to give complete answers.
	\item \emph{It can be estimated without bias for any finite $G$.} The per-sample Brier reward $r(y)-q(a(y))$ is linear in $q$. Estimating $q$ by the share of the same answer among the other samples, $\hat q_{-i}(a_i)=\frac1{G-1}\sum_{j\ne i}\one[a_j=a_i]$, gives $\E[\hat q_{-i}(a_i)\mid y_i]=q_\theta(a_i\mid x)$, so the unbaselined reward $R_i(\kappa)=r_i-(1-\kappa)\hat q_{-i}(a_i)$ is unbiased for any $G\ge2$. In contrast, $G$ samples can estimate without bias only polynomials in $q$ of degree at most $G-1$: the logarithmic score (with $1/p$) and power scores with non-integer exponent (with $q^{\gamma-1}$) cannot be estimated without bias for any finite $G$. Integer powers such as $\gamma=3$ can, but their reward to the correct answer, $\gamma p^{\gamma-2}(1-p)$, shrinks quickly as $p$ falls, so the push is weakest exactly when the correct answer is in the minority. $\gamma=2$ is the smallest estimable power and hence the friendliest to low-probability answers among the feasible choices.
	\item \emph{It belongs to the same family as pass@1.} At $\kappa=1$ it is exactly Dr.GRPO, so the second stage returns smoothly to the standard objective, which other proper scores do not allow.
\end{enumerate}

\textbf{Examples} (population advantages for large $G$).
\emph{L3 problem} (correct 0.15, most common wrong 0.50, seven other wrong answers at 0.05): Dr.GRPO gives $+0.85$ to the correct answer and $-0.15$ to every wrong one. PSPO's first stage, after subtracting the expected reward $-0.14$, gives $+0.99$ to the correct answer, $-0.36$ to the most common wrong answer (2.4 times the Dr.GRPO penalty) and $+0.09$ to the other wrong answers, redistributing the probability held by the common wrong answer.
\emph{L4 problem} (most common wrong 0.60, eight others at 0.05): Dr.GRPO and MaxRL give zero advantage everywhere; PSPO, after subtracting $-0.38$, gives $-0.22$ to the common wrong answer and $+0.33$ to the rare ones, so wrong answers no longer pile up in one place.
\emph{L2 problem} (correct 0.70, six wrong at 0.05): PSPO gives the correct answer only $+0.105$ (Dr.GRPO: $+0.30$), leaving more of the update to unsolved problems.

\subsection{Bias from subtracting the group mean, and a leave-one-out correction}
\label{sec:loo}

In practice the group mean is subtracted as a baseline to reduce variance. Doing so directly, $A_i=R_i-\frac1G\sum_jR_j$ as in GRPO and Dr.GRPO, breaks both unbiasedness and properness.

\begin{theorem}[Sharpening bias of the group mean and a leave-one-out correction]
\label{thm:loo}
Let $y_1,\dots,y_G\sim\pi_\theta(\cdot\mid x)$ be i.i.d., $g_i=\nabla_\theta\log\pi_\theta(y_i\mid x)$, $R_i(\kappa)=r_i-(1-\kappa)\hat q_{-i}(a_i)$, and let $\bar r$, $\bar{\hat q}$ and $\bar R(\kappa)$ be the group means of $r_j$, $\hat q_{-j}(a_j)$ and $R_j(\kappa)$.
\begin{enumerate}[leftmargin=*,itemsep=1pt,topsep=2pt]
	\item \emph{Subtracting the group mean leaves a sharpening term.} For any $G\ge2$,
	\begin{equation}
		\E\big[(R_i(\kappa)-\bar R(\kappa))\,g_i\big]=\tfrac{G-1}{G}\nabla_\theta q_\theta(a^*\mid x)-(1-\kappa)\tfrac{G-2}{G}\nabla_\theta\phiq(q_\theta).
		\label{eq:naive}
	\end{equation}
	At $\kappa=0$ this optimizes $\frac{G-1}{G}\big(S_{\brier}+\frac1{G-1}\phiq\big)$, i.e.\ a residual sharpening term of weight $\kappa_{\mathrm{eff}}=\frac1{G-1}$. For $G=2$ the Brier part cancels completely and the result equals Dr.GRPO; for $G=8$ the residual weight is still $1/7\approx14\%$.
	\item \emph{Leave-one-out correction.} For any $G\ge3$, use the baseline $\bar r_{-i}-(1-\kappa)\hat C_{-i}$, which depends only on the other samples, where $\bar r_{-i}=\frac1{G-1}\sum_{j\ne i}r_j$ and $\hat C_{-i}=\frac1{(G-1)(G-2)}\sum_{j\ne k;\,j,k\ne i}\one[a_j=a_k]$ is the unbiased U-statistic for the probability that two samples agree. The advantage simplifies to
	\begin{equation}
		A_i^{\mathrm{PSPO}}(\kappa)=\tfrac{G}{G-1}(r_i-\bar r)-(1-\kappa)\tfrac{G}{G-2}\big(\hat q_{-i}(a_i)-\bar{\hat q}\big),
		\label{eq:loo}
	\end{equation}
	which is unbiased for any finite $G\ge3$ and sums to zero within the group:
	\begin{equation}
		\E\big[A_i^{\mathrm{PSPO}}(\kappa)\,g_i\big]=\nabla_\theta\big(S_{\brier}+\kappa\,\phiq\big),\qquad\textstyle\sum_{i=1}^GA_i^{\mathrm{PSPO}}(\kappa)=0.
		\label{eq:loo_unbiased}
	\end{equation}
\end{enumerate}
\end{theorem}

The proof is in Appendix~\ref{app:loo}. The bias comes from the different orders of the two statistics. $r_i$ depends on one sample, so subtracting the group mean only rescales it by $\frac{G-1}{G}$. $\hat q_{-i}(a_i)$ depends on pairs of samples: $y_i$ appears not only in its own $\hat q_{-i}$ but also in the $\hat q_{-j}$ of the other $G-1$ samples, so subtracting the group mean shrinks it to $\frac{G-2}{G}$. Multiplying the centered $\hat q$ term by $\frac G{G-2}$ instead of $\frac G{G-1}$ removes the bias at no extra cost. For $G=2$ one can use the unbaselined $R_i(\kappa)$, or center only $r_i$.

\subsection{Two stages and cost}
\label{sec:twostage}

The $T_{\mathrm{tot}}$ training steps are split into two stages (Algorithm~\ref{alg:pspo}).
\emph{Stage 1: reranking and calibration ($\kappa=0$, about the first 80\% of steps)} uses the unbiased advantage $A_i^{\mathrm{PSPO}}(0)$ of Equation~\eqref{eq:loo}. It must not be divided by the group standard deviation, which would reweight problems by their accuracy and turn the objective back into an improper form like $2\arcsin\sqrt p$ (Section~\ref{sec:general}); to stabilize the scale, the whole batch may be divided by one global constant, which does not change the direction.
\emph{Stage 2: labeled sharpening ($\kappa$ from 0 to 1, about the last 20\% of steps).} Stage 1 yields a calibrated answer distribution: correct answers in the minority can become the mode and pass@$k$ rises clearly, but the distribution keeps some entropy and single-sample accuracy is still below mode decoding. Raising $\kappa$ linearly to 1 ends at unbiased Dr.GRPO. Stage 2 must keep the label term $r_i$ and cannot switch to a label-free sharpening reward: the ablation in Section~\ref{sec:exp} shows that when shared biases are learned faster than discriminative features, label-free sharpening destroys features already learned and lets the most common wrong answers take back the recovered L3 problems. A smaller learning rate or a KL penalty in Stage 2 also helps; if the model is deployed with majority voting, Stage 2 can be skipped.

\begin{algorithm}[t]
	\caption{Proper Scoring Policy Optimization (group size $G\ge3$)}
	\label{alg:pspo}
	\small
	\begin{algorithmic}[1]
		\Require policy $\pi_\theta$, training set $D$, group size $G\ge3$, steps $T_{\mathrm{tot}}$, stage-1 fraction $\rho$ (default 0.8)
		\For{$t=1,\dots,T_{\mathrm{tot}}$}
		\State $\kappa_t\gets 0$ if $t\le\rho T_{\mathrm{tot}}$, else $(t-\rho T_{\mathrm{tot}})/((1-\rho)T_{\mathrm{tot}})$
		\For{each problem $x$ in a batch from $D$}
		\State sample $y_1,\dots,y_G\sim\pi_\theta(\cdot\mid x)$; $a_i\gets$ equivalence class of $y_i$ ($\varnothing$ if no answer or cut)
		\State $r_i\gets\one[a_i=a^*(x)]$; $c(a)\gets$ number of samples with answer $a$
		\State $\hat q_i\gets(c(a_i)-1)/(G-1)$
		\State $A_i\gets\frac G{G-1}(r_i-\bar r)-(1-\kappa_t)\frac G{G-2}(\hat q_i-\bar{\hat q})$ \Comment{Eq.~\eqref{eq:loo}; no std normalization}
		\EndFor
		\State update $\theta$ with the clipped policy-gradient objective (token normalization as in Dr.GRPO)
		\EndFor
	\end{algorithmic}
\end{algorithm}

\textbf{Cost.} PSPO uses the same sampling as GRPO ($G$ responses per problem), the same number of verifier calls and the same forward and backward computation, with no teacher, process reward model or extra continuations. The only new computation is grouping the $G$ answers of a group into equivalence classes, which is negligible next to generation. PSPO can therefore replace the advantage computation of GRPO directly.

\section{Generalization: splitting GRPO and its variants}
\label{sec:general}

\subsection{Objectives that depend only on the probability of the correct answer}
\label{sec:local}

The split of Theorem~\ref{thm:identity} is not limited to pass@1. It applies to every objective $f(p)$ that depends only on the probability $p=q(a^*)$ of the correct answer, which the scoring-rule literature calls \emph{local} scores. Almost all online RL objectives built on binary outcomes are of this kind (see also~\cite{tajwar2026maxrl}): Dr.GRPO uses $f(p)=p$; GRPO divides by the group standard deviation, which in expectation corresponds to $f(p)=2\arcsin\sqrt p$~\cite{davis2025objective,tajwar2026maxrl}; pass@$k$ is $f(p)=1-(1-p)^k$; and MaxRL with group size $G$ estimates $f(p)=-\sum_{k=1}^{G-1}(1-p)^k/k$ without bias~\cite[Theorem~5]{tajwar2026maxrl}, whose limit as $G\to\infty$ is $\log p$.

\begin{theorem}[Unique split of local objectives]
\label{thm:local}
Let $f$ be differentiable and strictly increasing on $(0,1]$, and $F(t)=\int_0^tf(s)\,ds$. Then
\begin{equation}
	f\big(q(a^*)\big)=\underbrace{S_\phi(q,a^*)}_{\text{strictly proper, }\phi(q)=\sum_aF(q(a))}+\underbrace{\Psi_f(q)}_{\text{label-free, }\Psi_f(q)=\sum_a[q(a)f(q(a))-F(q(a))]},
	\label{eq:local}
\end{equation}
and the split is unique up to a constant under the requirement that the first part be strictly proper and the second be label-free. Per sample, the policy-gradient reward $f'(p)\,r(y)$ splits as
\begin{equation}
	f'(p)\,r(y)=\underbrace{f'\big(q(a)\big)\big[r(y)-q(a)\big]}_{\text{depends on the label}}+\underbrace{q(a)\,f'\big(q(a)\big)}_{\text{label-free}},\qquad a=a(y).
	\label{eq:local_sample}
\end{equation}
With at least three answers, the label-free part is constant if and only if $f(p)=c_1\log p+c_2$.
\end{theorem}

The proof is in Appendix~\ref{app:local}; $f(p)=p$ recovers Theorem~\ref{thm:identity}. Theorem~\ref{thm:local} gives a simple test: look at the label-free part $q\,f'(q)$ in Equation~\eqref{eq:local_sample}. It rewards each answer according to its current probability alone; if a higher probability earns a higher reward, probable answers become ever more probable, which is sharpening.
\begin{itemize}[leftmargin=*,itemsep=1pt,topsep=2pt]
	\item \emph{Dr.GRPO}: $q\,f'(q)=q$, increasing in $q$; it sharpens.
	\item \emph{GRPO}: $q\,f'(q)=\sqrt{q/(1-q)}$, increasing and unbounded as $q\to1$, so it pushes hardest on answers that already dominate.
	\item \emph{pass@$k$}: $q\,f'(q)=kq(1-q)^{k-1}$, increasing for $q<1/k$ and decreasing beyond: it sharpens among low-probability answers and flattens high-probability ones.
	\item \emph{MaxRL with group size $G$}: $q\,f'(q)=1-(1-q)^{G-1}$, increasing; it sharpens (Section~\ref{sec:maxrl} shows that it drives low-frequency answers to zero).
	\item \emph{Logarithmic score} (MaxRL as $G\to\infty$): $q\,f'(q)=1$, a constant with no effect on the update. It is the only proper local objective, in line with Bernardo~\cite{bernardo1979expected}, but its gradient contains $1/p$ and cannot be estimated without bias from finitely many samples.
\end{itemize}
Hence pass@1, GRPO, pass@$k$ and MaxRL with any finite group size are all improper, and they differ only in how strong their label-free part is and in which direction it pushes. For $b=(0.625,0.125,0.25)$, the optimal reports are $(1,0,0)$ for both pass@1 and GRPO, about $(0.73,0.01,0.26)$ for MaxRL with $G=8$, about $(0.51,0.16,0.33)$ for pass@4 (flatter than the truth), and exactly $b$ only for the logarithmic score (numerical checks in Section~\ref{sec:exp}).

\textbf{A common objective.} Why does the objective on the answer distribution suffice to tell these methods apart? The next proposition shows that within one answer all of them move in the same direction (proof in Appendix~\ref{app:within}).

\begin{proposition}[Within-answer and between-answer parts of the gradient]
\label{prop:within}
If the objective depends on the parameters only through the answer distribution, $J_x(\theta)=L(q_\theta(\cdot\mid x),a^*)$ with $L$ differentiable, then wherever $q_\theta(a\mid x)>0$,
\begin{equation}
	\nabla_\theta J_x(\theta)=\sum_{a\in\Acal}\frac{\partial L}{\partial q_\theta(a\mid x)}\,q_\theta(a\mid x)\,\E_{y\sim\pi_\theta(\cdot\mid x,\,a(y)=a)}\big[\nabla_\theta\log\pi_\theta(y\mid x)\big].
	\label{eq:within}
\end{equation}
\end{proposition}
Within one answer the update direction is always $\E[\nabla_\theta\log\pi_\theta(y\mid x)\mid a(y)=a]$. For the correct answer, Dr.GRPO, GRPO, MaxRL and PSPO move in exactly the same direction, which is why all of them ``amplify their own correct paths''. They differ only in the coefficients $\partial L/\partial q(a)$, that is, in how the update is distributed across answers. By Theorem~\ref{thm:local} these coefficients always split into a proper part and a label-free part, so labeled and label-free methods can be written as one objective,
\begin{equation}
	J_{\phi,\Psi;\alpha,\beta}(\theta)=\alpha\,\E_x\big[S_\phi(q_\theta(\cdot\mid x),a^*)\big]+\beta\,\E_x\big[\Psi(q_\theta(\cdot\mid x))\big],\qquad\alpha,\beta\ge0,
	\label{eq:unified}
\end{equation}
where $S_\phi$ is a proper score and $\Psi$ any label-free function. This general objective is a tool for analyzing existing methods, not our method. Local objectives have $\alpha=\beta=1$ with $(\phi,\Psi)$ from Theorem~\ref{thm:local}, and Dr.GRPO is the special case $\Psi=\phi$; label-free methods have $\alpha=0$. PSPO (Section~\ref{sec:method}) is a single point in this family: the Brier score with the sharpening term removed ($\phi=\phiq$, $\alpha=1$, $\beta=0$), with $\beta=\kappa$ raised only at the end of training.

\begin{theorem}[Policy gradient of the common objective]
\label{thm:unified}
If $\phi$ and $\Psi$ are twice continuously differentiable in the interior of the simplex, then
\begin{equation}
	\begin{gathered}
	\nabla_\theta J_{\phi,\Psi;\alpha,\beta}(\theta)=\E_{x,\,y\sim\pi_\theta(\cdot\mid x)}\big[R(y)\nabla_\theta\log\pi_\theta(y\mid x)\big],\\
	R(y)=\alpha\big[\nabla^2\phi(q)(e_{a^*}-q)\big]_{a(y)}+\beta\big[\nabla\Psi(q)\big]_{a(y)}.
	\end{gathered}
	\label{eq:unified_grad}
\end{equation}
\end{theorem}
The proof is in Appendix~\ref{app:unified}. The first term of $R(y)$ reranks answers using the label, and the second sharpens (or flattens) without it; for local objectives they are the two terms of Equation~\eqref{eq:local_sample}. Table~\ref{tab:methods} lists the split of each method.

\begin{table}[t]
	\centering
	\caption{Methods under the common objective (Equation~\ref{eq:unified}). $r=r(y)$, $q(a)=q(a(y))$, $p=q(a^*)$. For local objectives the two reward columns sum to $f'(p)\,r$; GRPO's actual advantage $(r-p)/\sqrt{p(1-p)}$ is a baselined version with the same expected gradient.}
	\label{tab:methods}
	\small
	\setlength{\tabcolsep}{3.5pt}
	\resizebox{\linewidth}{!}{%
	\begin{tabular}{lllll}
		\toprule
		Method & Objective & Proper part (per sample) & Label-free part (per sample) & Effect \\
		\midrule
		\multicolumn{5}{l}{\emph{Local objectives ($\alpha=\beta=1$, Theorem~\ref{thm:local})}} \\
		Dr.GRPO & $p$ & $r-q(a)$ & $q(a)$ & removes L1/L2; hurts L3 \\
		GRPO & $2\arcsin\sqrt p$ & $\frac{r-q(a)}{\sqrt{q(a)(1-q(a))}}$ & $\sqrt{\frac{q(a)}{1-q(a)}}$ & stronger sharpening \\
		pass@$k$ & $1-(1-p)^k$ & $k(1-q(a))^{k-1}[r-q(a)]$ & $kq(a)(1-q(a))^{k-1}$ & sharpens low, flattens high \\
		MaxRL ($G$) & $-\sum_{k<G}\frac{(1-p)^k}{k}$ & $\frac{1-(1-q(a))^{G-1}}{q(a)}[r-q(a)]$ & $1-(1-q(a))^{G-1}$ & zeroes rare answers (Thm.~\ref{thm:maxrl}) \\
		MaxRL ($G\to\infty$) & $\log p$ & $r/p-1$ & constant & proper; not estimable at finite $G$ \\
		\midrule
		\multicolumn{5}{l}{\emph{Ours: PSPO, the Brier score with the sharpening term removed (Section~\ref{sec:method})}} \\
		PSPO stage 1 & -- & $r-q(a)$ & $0$ & proper; unbiased at finite $G$ \\
		PSPO stage 2 & -- & $r-q(a)$ & $\kappa\,q(a)$, $\kappa:0\to1$ & labeled sharpening \\
		\midrule
		\multicolumn{5}{l}{\emph{Label-free methods ($\alpha=0$)}} \\
		soft majority vote / TTRL & -- & none & $q(a)$ & sharpening only \\
		answer-level entropy min. & -- & none & $\log q(a)$ & sharpening only \\
		\bottomrule
	\end{tabular}}
\end{table}

This framework also gives a simple check for new training objectives: write down the label-free part and see how it varies with the probability of an answer. If it increases, the objective sharpens; if it decreases, it flattens; only when it is constant does the objective leave the true distribution in place.

\subsection{MaxRL with a finite group size}
\label{sec:maxrl}

Theorem~\ref{thm:local} shows that MaxRL with a finite group size still contains a sharpening part. The next result gives its consequence (proof in Appendix~\ref{app:maxrl}).

\begin{theorem}[MaxRL with a finite group suppresses rare answers]
\label{thm:maxrl}
Let $b$ have at least two answers with positive probability and maximize $\E_{a^*\sim b}[J^{(G-1)}_{\mathrm{MaxRL}}(q(a^*))]$ over a shared $q$, where $J^{(G-1)}_{\mathrm{MaxRL}}(p)=-\sum_{k=1}^{G-1}(1-p)^k/k$. At $q=b$,
\begin{equation}
	\frac{\partial}{\partial q(a)}\E_b\big[J^{(G-1)}_{\mathrm{MaxRL}}(q(a^*))\big]\Big|_{q=b}=1-\big(1-b(a)\big)^{G-1},
	\label{eq:maxrl}
\end{equation}
which strictly increases with $b(a)$, so $q=b$ is not optimal unless all answers with positive probability are equally likely. The optimum $q^*$ favors frequent answers and sets to exactly zero every answer with $b(a)\le\lambda^*/(G-1)$, where $\lambda^*\in(0,1]$ is the Lagrange multiplier of $\sum_aq(a)=1$.
\end{theorem}

To keep an answer of true frequency $b_{\min}$, MaxRL therefore needs a group size $G=\Omega(1/b_{\min})$. For $b=[0.65,0.20,0.08,0.05,0.02]$, $G=8$ drives the last three answers below $3\times10^{-4}$ and $G=16$ still drives the last two below $8\times10^{-4}$ (Section~\ref{sec:exp}). Even as $G\to\infty$, the logarithmic score as a local score has two problems. First, it gives every wrong sample the same score and does not distinguish between wrong answers, while the key to L3 is to demote the single most common wrong answer: on m300, in 41.4\%--93.7\% of the problems with wrong samples the most common wrong answer accounts for more than half of the wrong samples (36.3\%--68.1\% for five models after forcing R1 to answer and removing empty answers). Second, when all samples of a group are wrong (L4), every sample gets the same score and the baselined gradient is zero.

\section{Simulation experiments}
\label{sec:exp}

Before training large models for long horizons, we test the theory, and training under shared parameters, in controlled linear-softmax simulations (code and full outputs in the supplementary material).

\subsection{Numerical checks}
\label{sec:checks}

\emph{Finite differences.} For random parameters $z\in\mathbb{R}^5$, the Brier gradient has maximum absolute error $2.3\times10^{-11}$, the pass@1 identity has error 0, and the power-score reward ($\gamma=1.5$) has error $7.4\times10^{-11}$. For pass@1, GRPO, pass@4, MaxRL with $G=8$ and the logarithmic score, the split of Theorem~\ref{thm:local} holds on random distributions with error at most $1.3\times10^{-15}$, the gradient of the label-free part matches $q\,f'(q)$ within $1.1\times10^{-10}$, and the optimum of the proper part matches the true distribution within the grid resolution (0.001).

\emph{Finite-$G$ expectations} ($4\times10^5$ Monte Carlo samples, Theorem~\ref{thm:loo}). Subtracting the group mean, $R_i-\bar R$, matches Equation~\eqref{eq:naive} for $G\in\{2,3,4,8\}$ (maximum error $4\times10^{-5}$ to $2.6\times10^{-4}$) and deviates clearly from the true Brier gradient (after rescaling by $\frac G{G-1}$, errors 0.0345, 0.0172, 0.0115 and 0.0049, proportional to $\frac1{G-1}$); the leave-one-out correction of Equation~\eqref{eq:loo} matches the true Brier gradient for $G\in\{3,4,8\}$ up to Monte Carlo noise ($4.1\times10^{-4}$, $2.9\times10^{-4}$, $4\times10^{-5}$).

\emph{Optimum of MaxRL with finite $G$} (Theorem~\ref{thm:maxrl}). For $b=[0.65,0.20,0.08,0.05,0.02]$, the maximizer of $J^{(G-1)}_{\mathrm{MaxRL}}$ is $[0.8068,0.1926,0.0003,0.0002,0.0001]$ for $G=8$, $[0.7338,0.2203,0.0449,0.0008,0.0002]$ for $G=16$, and only for $G=64$ roughly $b$: $[0.6595,0.2029,0.0808,0.0485,0.0083]$.

\subsection{Three environments with shared parameters}
\label{sec:envs}

\textbf{Setup.} Each problem has $K=5$ candidate answers, answer 0 being the ``usual method''. Problems fall into $C=20$ classes, and in each class the usual method is correct for a fraction $\rho\in[0.45,0.85]$ of the problems. The logits are $z=W_c+Uh$. The class bias $W_c$ starts at $[1.6,0.3,-1.0,-2.0,-3.0]$, an initial distribution of about $[0.72,0.20,0.05,0.02,0.007]$ that contains L2, L3 and L4 problems at once; it represents the fast, shared sharpening direction. The feature term $Uh$ ($D=8$, initialized at zero) represents features that separate special problems and have to be learned gradually. Training and test sets have 4,000 problems each; every step uses 64 problems with $G=8$ samples and Adam (learning rate 0.02 for $W$); every method trains for 2,000 steps, and results average three seeds.
\emph{Environment A (indistinguishable classes):} whether the correct answer departs from the usual method is unrelated to $h$; it tests calibration and whether low-probability answers survive.
\emph{Environment B (easy features):} $h$ indicates the correct answer with noise, and $U$ and $W$ share a learning rate.
\emph{Environment C (slow features):} as B, but $U$ learns at 1/10 of the rate of $W$, the case where sharpening is much faster than learning new discriminative features, as in large models.
Besides pass@1 at $T=1.0$ and $T=0.6$, mode-decoding accuracy (argmax, the ceiling of single-sample accuracy after complete sharpening) and pass@16, we track each test problem with its exact probabilities (L4: $p<0.01$; L3: $p\ge0.01$ and not the mode; L2: the mode with $p<0.9$; solved: the mode with $p\ge0.9$). Table~\ref{tab:toy} reports the results.

\begin{table}[t]
	\centering
	\caption{Test results in the three simulated environments (4,000 test problems, 2,000 training steps, mean of three seeds). p@1: single-sample accuracy at the given temperature; argmax: mode-decoding accuracy; H: answer entropy. Transition columns are fractions of the initial layer: L2$\to$L3/4, L2 problems whose correct answer is no longer the mode; L3$\to$L4, L3 problems whose correct-answer probability falls below 0.01; L3$\to$L2/s, L3 problems whose correct answer becomes the mode; L4$\to$\,$\neg$L4, L4 problems whose correct-answer probability rises above 0.01.}
	\label{tab:toy}
	\footnotesize
	\setlength{\tabcolsep}{3.2pt}
	\resizebox{\linewidth}{!}{%
	\begin{tabular}{lccccccccc}
		\toprule
		Method & p@1 (1.0) & p@1 (0.6) & argmax & pass@16 & H & L2$\to$L3/4 & L3$\to$L4 & L3$\to$L2/s & L4$\to$\,$\neg$L4 \\
		\midrule
		\multicolumn{10}{l}{\emph{Environment A, indistinguishable classes} (initial 0.490 / 0.580 / 0.644 / 0.817; seed 0: L2 0.65, L3 0.25, L4 0.09)} \\
		GRPO & 0.644 & 0.644 & 0.644 & 0.645 & 0.01 & 0.00 & 1.00 & 0.00 & 0.00 \\
		Dr.GRPO & 0.644 & 0.644 & 0.644 & 0.645 & 0.01 & 0.00 & 1.00 & 0.00 & 0.00 \\
		GRPO + entropy (0.05) & 0.643 & 0.644 & 0.644 & 0.646 & 0.01 & 0.00 & 1.00 & 0.00 & 0.00 \\
		MaxRL ($G=8$) & 0.605 & 0.626 & 0.637 & 0.718 & 0.28 & 0.01 & 0.59 & 0.01 & 0.22 \\
		Brier, group-mean baseline & 0.515 & 0.594 & 0.643 & 0.858 & 0.84 & 0.00 & 0.04 & 0.00 & 0.75 \\
		\textbf{PSPO stage 1} & 0.466 & 0.566 & 0.643 & \textbf{0.903} & 1.06 & 0.00 & \textbf{0.00} & 0.00 & \textbf{0.93} \\
		PSPO two-stage (1600+400) & 0.623 & 0.642 & 0.644 & 0.708 & 0.23 & 0.00 & 0.45 & 0.00 & 0.40 \\
		\emph{stage 2 without labels} & 0.640 & 0.644 & 0.644 & 0.654 & 0.05 & 0.00 & 1.00 & 0.00 & 0.00 \\
		\midrule
		\multicolumn{10}{l}{\emph{Environment B, easy features} (initial 0.490 / 0.578 / 0.640 / 0.831; seed 0: L2 0.64, L3 0.33, L4 0.03)} \\
		GRPO & 0.806 & 0.815 & 0.822 & 0.915 & 0.17 & 0.08 & 0.09 & 0.70 & 0.60 \\
		Dr.GRPO & 0.805 & 0.815 & 0.821 & 0.913 & 0.18 & 0.08 & 0.10 & 0.69 & 0.56 \\
		GRPO + entropy (0.05) & 0.807 & 0.817 & 0.825 & 0.922 & 0.19 & 0.08 & 0.08 & 0.70 & 0.76 \\
		MaxRL ($G=8$) & 0.794 & 0.812 & 0.823 & 0.944 & 0.27 & 0.09 & 0.04 & 0.69 & 0.86 \\
		Brier, group-mean baseline & 0.780 & 0.807 & 0.825 & 0.954 & 0.33 & 0.09 & 0.02 & 0.70 & 0.88 \\
		\textbf{PSPO stage 1} & 0.762 & 0.798 & 0.824 & \textbf{0.965} & 0.42 & 0.09 & \textbf{0.01} & 0.70 & \textbf{0.92} \\
		PSPO two-stage (1600+400) & 0.785 & 0.807 & 0.820 & 0.949 & 0.31 & 0.09 & 0.03 & 0.70 & 0.87 \\
		\emph{stage 2 without labels} & 0.751 & 0.758 & 0.761 & 0.832 & 0.15 & 0.12 & 0.23 & 0.65 & 0.10 \\
		\midrule
		\multicolumn{10}{l}{\emph{Environment C, slow features} (initial distribution as in B; $U$ learns at 1/10 of the rate of $W$)} \\
		GRPO & 0.769 & 0.781 & 0.787 & 0.866 & 0.20 & 0.04 & 0.21 & 0.56 & 0.08 \\
		Dr.GRPO & 0.768 & 0.780 & 0.787 & 0.860 & 0.19 & 0.04 & 0.25 & 0.56 & 0.02 \\
		GRPO + entropy (0.05) & 0.776 & 0.793 & 0.806 & 0.908 & 0.26 & 0.05 & 0.11 & 0.60 & 0.32 \\
		MaxRL ($G=8$) & 0.759 & 0.794 & 0.816 & 0.961 & 0.41 & 0.07 & 0.02 & 0.66 & 0.80 \\
		Brier, group-mean baseline & 0.742 & 0.787 & 0.819 & 0.971 & 0.49 & 0.07 & 0.01 & 0.67 & 0.94 \\
		\textbf{PSPO stage 1} & 0.723 & 0.779 & \textbf{0.821} & \textbf{0.978} & 0.57 & 0.08 & \textbf{0.00} & \textbf{0.67} & \textbf{0.96} \\
		\textbf{PSPO two-stage (1600+400)} & 0.754 & \textbf{0.788} & 0.808 & 0.958 & 0.42 & 0.05 & 0.02 & 0.60 & 0.91 \\
		\emph{stage 2 without labels} & 0.712 & 0.714 & 0.712 & 0.835 & 0.17 & 0.00 & 0.22 & 0.25 & 0.37 \\
		\bottomrule
	\end{tabular}}
\end{table}

The results agree with the theory in five respects.

\textbf{Environment A confirms Theorems~\ref{thm:maxrl} and~\ref{thm:loo}.} GRPO, Dr.GRPO and untuned GRPO with an entropy bonus move all L3 problems into L4, and pass@16 falls to 0.645, equal to pass@1. MaxRL with $G=8$, whose truncated objective is improper, still moves 59\% of L3 problems into L4 (pass@16 0.718), and the Brier score with a group-mean baseline, which keeps a sharpening term of weight $1/7$, still moves 4\% (pass@16 0.858). With the leave-one-out correction, PSPO stage 1 has L3$\to$L4 of 0\%, L4$\to\neg$L4 of 93\% and pass@16 of 0.903, which is 25.8, 18.5 and 4.5 points above GRPO, MaxRL and the group-mean Brier score.

\textbf{When PSPO raises the ceiling of single-sample accuracy.} Environment A has no learnable features, and argmax is 0.643--0.644 for every method. In environment B the features are learned before GRPO suppresses the minority answers, so the two reach the same argmax (0.824 vs.\ 0.822), and the advantage of the proper score appears in pass@16 (0.965 vs.\ 0.915) and L4$\to\neg$L4 (92\% vs.\ 60\%). The difference appears in environment C, where features are learned more slowly than sharpening. GRPO and Dr.GRPO first move 21\%--25\% of L3 problems into L4 through the fast bias $W$; these problems no longer produce correct samples, so the features $U$ receive no gradient for them and argmax stalls at 0.787 (L3$\to$L2/s 56\%, L4$\to\neg$L4 2\%--8\%). PSPO stage 1 never moves an L3 problem into L4 (L3$\to$L4 0\%), reaches L3$\to$L2/s of 67\%, L4$\to\neg$L4 of 96\% and pass@16 of 0.978 (11.2 points higher), and an argmax of 0.821, 3.4 points above GRPO and Dr.GRPO (MaxRL: 0.816).

\textbf{Two-stage PSPO.} With the same 2,000 steps, the first 1,600 use stage 1 and the last 400 raise $\kappa$ linearly from 0 to 1. In environment C, two-stage PSPO is slightly more accurate than GRPO and Dr.GRPO at $T=0.6$ (0.788 vs.\ 0.781 and 0.780) while keeping pass@16 9.2 points higher (0.958 vs.\ 0.866), with L3$\to$L4 of only 2\% and L4$\to\neg$L4 of 91\%.

\textbf{Stage 2 must keep the labels.} In environment C, raising $\kappa$ at the full learning rate lets the bias $W$, still ten times faster than $U$, pull argmax back from 0.821 to 0.808 (L3$\to$L2/s from 67\% to 60\%). In environment A, which has no learnable features, stage 2 trades pass@16 for single-sample accuracy: p@1 at $T=1.0$ rises from 0.466 to 0.623, but L3$\to$L4 returns to 45\% and pass@16 falls to 0.708. If stage 2 drops the labels and sharpens without them, argmax in environment C falls to 0.712, p@1 at $T=0.6$ to 0.714 and L3$\to$L2/s to 25\%. Stage 2 must therefore be labeled, gradual sharpening, and in large models it should come with a smaller learning rate or a KL penalty.

\textbf{Costs.} In environments B and C, the proper score lets slightly more initial L2 problems lose their mode (environment C: 8\% for stage 1, 5\% for two-stage PSPO, 4\% for GRPO). Two-stage PSPO keeps more entropy, so its single-sample accuracy at $T=1.0$ is below GRPO's in all three environments (environment C: 0.754 vs.\ 0.769), and in environment B it is also 0.8 points lower at $T=0.6$ (0.807 vs.\ 0.815). Overall the gains clearly outweigh the costs.

\section{Discussion, limitations and conclusion}
\label{sec:discussion}

\subsection{The sharpening debate and related work}
\label{sec:related}

Back to the question of the introduction: does online RL sharpen existing habits or learn new abilities? Our answer draws the line between L3 and L4. Online RL is not bound to sharpen: GRPO and Dr.GRPO remove L1 and L2 errors while hurting L3 not because they learn only from the model's own samples, but because they use an improper score on the answer distribution, which adds a fixed label-free sharpening reward on top of reranking. A proper score that is unbiased at finite $G$, as in PSPO, demotes the most common wrong answer, keeps correct answers in the minority, and goes beyond sharpening when discriminative features are learned slowly. Pure online RL, however, cannot obtain abilities outside its sampling range: for L4 problems with $q_\theta(a^*\mid x)=0$, every label-dependent part of an on-policy score-function gradient is zero. On all-wrong groups PSPO encourages exploration by demoting the most common wrong answer, but this exploration has no direction; targeted progress on L4 still requires offline distillation or external search.

\textbf{MaxRL}~\cite{tajwar2026maxrl} starts from maximum likelihood and the harmonic expansion of pass@$k$ and points out the limits of expected-reward objectives; its $G\to\infty$ limit is the logarithmic score, the only local objective without a label-free part (Theorem~\ref{thm:local}). We add that its truncated objective at finite $G$ is still improper and drives answers with frequency below about $1/(G-1)$ to zero, and that the non-local Brier score uses the differences between wrong answers, is unbiased for any $G\ge3$, and keeps a gradient on all-wrong groups.

\textbf{Outcome-based exploration}~\cite{song2025obe} uses history-based UCB bonuses and a batch penalty on repeated outputs to counter the loss of diversity, a form close to a negative reward on repeated answers, added to GRPO's standardized advantage with a tuned coefficient. Under our split, this penalty works by cancelling the sharpening term. Proper scoring further shows that (i) relative to the 0/1 reward, the coefficient of the leave-one-out share $-\hat q_{-i}(a_i)$ must be exactly 1, the only strictly proper choice in the quadratic family (smaller leaves sharpening, larger over-flattens); (ii) this is exactly the removal of the sharpening term from pass@1; (iii) the advantage should not be divided by the group standard deviation, which reweights problems by their accuracy and makes the objective improper again, whereas a penalty added after standardization has a weight relative to the 0/1 reward that varies across problems; (iv) when the whole reward is centered by the group mean, the $\hat q$ term must be scaled by an extra $\frac{G-1}{G-2}$ to stay unbiased; and (v) the same split into a proper part and a label-free part applies to any strictly convex $\phi$, which is how Section~\ref{sec:general} analyzes other objectives.

\textbf{Label-free methods} such as TTRL~\cite{zuo2025ttrl} use the majority answer as a pseudo-label; their smoothed version is exactly the sharpening term (Table~\ref{tab:methods}), so they remove only L1 and L2 errors and hurt L3. \textbf{RLCR}~\cite{damani2025rlcr} trains the confidence a model states in words with proper scores such as Brier; PSPO applies a proper score to the answer distribution obtained by sampling. The two calibrate different objects and are complementary. \textbf{On-policy self-distillation}~\cite{zhao2026opsd} uses a frozen copy of the model, given a reference solution in its context, as the teacher, and thus uses more information than a 0/1 verifier. If the teacher's extra information is only the final answer, sampling-based token-level reverse KL tends to concentrate on the mode and, under shared representations, yields an over-concentrated distribution, which is again improper.

\subsection{Limitations}
\label{sec:limits}

\begin{enumerate}[leftmargin=*,itemsep=1pt,topsep=2pt]
	\item \emph{A proper score adds no new information.} By Proposition~\ref{prop:bregman}, a proper score drives the calibration error $D_\phi(b\|q)$ to zero but cannot lower $H_\phi(A^*\mid Z)$, which is set by the representation. If an L3 problem is wrong because the representation cannot separate it, PSPO keeps its pass@$k$ but cannot make it correct in a single sample through the score alone.
	\item \emph{Answers must be groupable.} The leave-one-out share needs the responses of a group to fall reliably into discrete answers. Mathematical problems with final answers, code with unit tests and multiple-choice questions satisfy this; open-ended proofs or free text need a semantic equivalence check, for example by a model.
	\item \emph{Stage 2 is sensitive to its schedule.} Section~\ref{sec:exp} shows that annealing too fast, a learning rate that is too high, or a label-free sharpening reward lets the fast biases partly undo stage 1. The schedule, learning rate and KL strength on real models need systematic ablation.
	\item \emph{Short training or easy test sets may hide the difference.} The measured 150-step GRPO run (pass@16 unchanged on MATH-500, down 3 points on AMC) and the simulations both indicate that if training has not yet entered the phase where L3 problems fall into L4, or the test set consists mostly of saturated L2 problems, the advantage of PSPO can be masked by averages; per-problem transitions, hard competition problems and longer training are needed.
	\item \emph{Validation on large models is in progress.} The simulations set by construction that features are learned more slowly than biases; whether large models are closer to environment B or C has to be decided by training them, which we leave to an updated version of this paper.
\end{enumerate}

\subsection{Conclusion}
\label{sec:conclusion}

Online RL should not be judged by average single-sample accuracy alone, but by the four layers of the single-sample error: temperature noise, whether the correct answer is the mode, whether correct answers in the minority survive, and whether sampling covers the correct answer at all. Viewing the answer distribution as a probabilistic prediction shows that pass@1 is the Brier score plus a label-free sharpening term, which explains why GRPO improves the problems whose correct answer is already in the majority while suppressing correct answers in the minority. Removing this term, and correcting the bias of the group-mean baseline with a leave-one-out estimator, lets PSPO handle problems whose correct answer is in the minority, with no extra samples and no teacher. The same split covers GRPO, pass@$k$ and MaxRL, and gives a computable, testable boundary between what pure online learning can achieve and what requires distillation or external search.

{\small
\bibliographystyle{ieeenat_fullname}
\bibliography{main}
}

\clearpage
\appendix

\section{Proofs}
\label{app:proofs}

\subsection{Proof of Proposition~\ref{prop:bregman}}
\label{app:bregman}

Setting $u=e_{a^*}$ and $v=q$ in the definition of the Bregman divergence and comparing with Savage's representation gives $\phi(e_{a^*})-S_\phi(q,a^*)=D_\phi(e_{a^*}\|q)$. Taking the expectation over $a^*\sim b$ and using $\E_{a^*\sim b}[e_{a^*}]=b$,
\begin{equation*}
	\begin{aligned}
	\E_{a^*\sim b}\big[\phi(e_{a^*})-S_\phi(q,a^*)\big]&=\E_b[\phi(e_{a^*})]-\phi(q)-\langle\nabla\phi(q),b-q\rangle\\
	&=\big(\E_b[\phi(e_{a^*})]-\phi(b)\big)+\big(\phi(b)-\phi(q)-\langle\nabla\phi(q),b-q\rangle\big).
	\end{aligned}
\end{equation*}
Since $\phi$ is strictly convex, the second term $D_\phi(b\|q)\ge0$ with equality if and only if $q=b$; by Jensen's inequality the first term $H_\phi(A^*\mid Z)\ge0$, and it does not depend on $q$. \hfill$\blacksquare$

\subsection{Proof of Theorem~\ref{thm:identity}}
\label{app:identity}

\emph{Optimum of pass@1.} $\max_{q\in\Delta(\Acal)}\sum_ab(a)q(a)=\max_ab(a)$; if the mode $a_0$ is unique, the maximum is attained only at $q=\delta_{a_0}$.

\emph{Identity and gradient.} $\tfrac12\brier(q,a^*)+\tfrac12\sum_aq(a)^2=\tfrac12\big(2q(a^*)-\sum_aq(a)^2\big)+\tfrac12\sum_aq(a)^2=q(a^*)$. Differentiating, and using $\nabla_\theta q_\theta(a)=\E_{y\sim\pi_\theta}[\one[a(y)=a]\nabla_\theta\log\pi_\theta(y\mid x)]$ and
\begin{equation*}
	\nabla_\theta\Big(\tfrac12\textstyle\sum_aq_\theta(a)^2\Big)=\textstyle\sum_aq_\theta(a)\nabla_\theta q_\theta(a)=\E_{y\sim\pi_\theta}\big[q_\theta(a(y)\mid x)\nabla_\theta\log\pi_\theta(y\mid x)\big],
\end{equation*}
gives Equation~\eqref{eq:identity_grad}. $\tfrac12\brier=S_{\phiq}$ is strictly proper: its expectation $\sum_ab(a)q(a)-\tfrac12\sum_aq(a)^2$ is strictly concave, and the first-order condition $b(a)-q(a)=\text{const}$ under $\sum_aq(a)=1$ has the unique solution $q=b$.

\emph{Class-level split (Equation~\ref{eq:class}).} Rearrange $\|q-b\|_2^2=\sum_aq(a)^2-2\sum_ab(a)q(a)+\sum_ab(a)^2$.

\emph{Uniqueness.} Let $X$ be differentiable and depend only on $q$. If $q(a^*)-X(q)$ is strictly proper, then for every $b$ in the interior of the simplex, $\sum_ab(a)q(a)-X(q)$ is maximized over the simplex at $q=b$, and the Lagrange condition gives $\nabla X(b)=b+\lambda(b)\mathbf 1$. Since $\mathbf 1$ is orthogonal to the tangent directions of the simplex, $X$ and $\tfrac12\sum_aq(a)^2$ have the same derivative along every direction in the simplex and differ by a constant. This is also the case $f(p)=p$ of the uniqueness in Theorem~\ref{thm:local} (Appendix~\ref{app:local}). \hfill$\blacksquare$

\subsection{Proof of Theorem~\ref{thm:loo}}
\label{app:loo}

Fix $x$ and i.i.d.\ samples $y_1,\dots,y_G\sim\pi_\theta(\cdot\mid x)$ with $g_i=\nabla_\theta\log\pi_\theta(y_i\mid x)$. Then $\E[g_i]=0$, $\E[r_ig_i]=\nabla_\theta q_\theta(a^*)$, and for any $j\ne i$,
\begin{equation*}
	\E\big[\one[a_j=a_i]\,g_i\big]=\E_{y_i}\big[\Pr(a_j=a_i\mid y_i)\,g_i\big]=\E_{y_i}\big[q_\theta(a_i)\,g_i\big]=\nabla_\theta\phiq(q_\theta).
\end{equation*}

\emph{Part 1 (bias of the group mean).} By independence, $\E[r_jg_i]=\E[r_j]\E[g_i]=0$ for $j\ne i$, so $\E[(r_i-\bar r)g_i]=\frac{G-1}G\nabla_\theta q_\theta(a^*)$. For the products of $\hat q_{-j}(a_j)$ and $g_i$: if $j=i$, $\E[\hat q_{-i}(a_i)g_i]=\frac1{G-1}\sum_{k\ne i}\E[\one[a_k=a_i]g_i]=\nabla_\theta\phiq$; if $j\ne i$, the only term of $\hat q_{-j}(a_j)=\frac1{G-1}\one[a_i=a_j]+\frac1{G-1}\sum_{k\notin\{i,j\}}\one[a_k=a_j]$ that contains $y_i$ is the first, which contributes $\frac1{G-1}\nabla_\theta\phiq$. Averaging over $j$,
\begin{equation*}
	\E[\bar{\hat q}\,g_i]=\tfrac1G\nabla_\theta\phiq+\tfrac{G-1}G\cdot\tfrac1{G-1}\nabla_\theta\phiq=\tfrac2G\nabla_\theta\phiq,
\end{equation*}
so $\E[(\hat q_{-i}(a_i)-\bar{\hat q})g_i]=\frac{G-2}G\nabla_\theta\phiq$. Combining with the $r$ term gives Equation~\eqref{eq:naive}.

\emph{Part 2 (leave-one-out correction).} For $G\ge3$, $\bar r_{-i}$ and $\hat C_{-i}$ depend only on $\{y_j\}_{j\ne i}$ and are independent of $y_i$, so $\E[\bar r_{-i}g_i]=\E[\hat C_{-i}g_i]=0$ and
\begin{equation*}
	\E\Big[\big(r_i-\bar r_{-i}-(1-\kappa)(\hat q_{-i}(a_i)-\hat C_{-i})\big)g_i\Big]=\nabla_\theta q_\theta(a^*)-(1-\kappa)\nabla_\theta\phiq=\nabla_\theta\big(S_{\brier}+\kappa\,\phiq\big).
\end{equation*}
For the closed form, $r_i-\bar r_{-i}=r_i-\frac{G\bar r-r_i}{G-1}=\frac G{G-1}(r_i-\bar r)$. The number of ordered equal pairs in the group is $C=\sum_{j\ne k}\one[a_j=a_k]=G(G-1)\bar{\hat q}$, of which $2(c(a_i)-1)=2(G-1)\hat q_{-i}(a_i)$ involve sample $i$, so
\begin{equation*}
	\hat C_{-i}=\frac{C-2(G-1)\hat q_{-i}(a_i)}{(G-1)(G-2)}=\frac{G\bar{\hat q}-2\hat q_{-i}(a_i)}{G-2},
\end{equation*}
hence
\begin{equation*}
	\hat q_{-i}(a_i)-\hat C_{-i}=\frac G{G-2}\big(\hat q_{-i}(a_i)-\bar{\hat q}\big).
\end{equation*}
Substituting gives Equation~\eqref{eq:loo}; the sum over $i$ is clearly zero. \hfill$\blacksquare$

\subsection{Proof of Theorem~\ref{thm:local}}
\label{app:local}

\emph{The split holds.} With $\phi(q)=\sum_aF(q(a))$, $\partial\phi/\partial q(a)=f(q(a))$, and Savage's representation gives $S_\phi(q,a^*)=\sum_aF(q(a))+f(q(a^*))-\sum_aq(a)f(q(a))$, so $S_\phi+\Psi_f=f(q(a^*))$, which is Equation~\eqref{eq:local}. Since $F''=f'>0$, $\phi$ is a sum of strictly convex functions and hence strictly convex, so $S_\phi$ is strictly proper; $\Psi_f$ depends only on $q$.

\emph{Per-sample form.} By Theorem~\ref{thm:unified} with $\alpha=\beta=1$ and $\Psi=\Psi_f$, $\nabla^2\phi(q)=\mathrm{diag}(f'(q(a)))$, so the label-dependent part is $f'(q(a))[\one[a=a^*]-q(a)]$, and $\partial\Psi_f/\partial q(a)=f(q(a))+q(a)f'(q(a))-f(q(a))=q(a)f'(q(a))$. Their sum is $f'(p)(1-p)+pf'(p)=f'(p)$ for $a=a^*$ and 0 otherwise, i.e.\ $f'(p)\,r$, which is Equation~\eqref{eq:local_sample}.

\emph{Uniqueness.} Let $f(q(a^*))=S(q,a^*)+X(q)$ with $S$ strictly proper (with differentiable generating function $\psi$) and $X$ label-free. By Savage's representation, $S(q,a^*)=[\psi(q)-\langle\nabla\psi(q),q\rangle]+\partial\psi/\partial q(a^*)$, where only the last term depends on the label. Comparing the label-dependent parts, $\partial\psi/\partial q(a)-f(q(a))$ is the same for all $a$, i.e.\ $\nabla\psi(q)=(f(q(a)))_a+c(q)\mathbf 1$. As $\mathbf 1$ is orthogonal to the simplex, $\psi$ and $\phi=\sum_aF(q(a))$ have the same derivative along every direction in it and differ by a constant, so $S=S_\phi$ and $X=\Psi_f$ up to constants.

\emph{When the label-free part vanishes.} $\Psi_f$ is constant on the simplex if and only if its gradient has equal components everywhere, i.e.\ $q(a)f'(q(a))$ does not depend on $a$ for any $q$. With at least three answers, any $s,t\in(0,1)$ with $s+t<1$ can be two components of one distribution, so $t\,f'(t)$ is a constant $c_1$ on $(0,1)$, i.e.\ $f(p)=c_1\log p+c_2$. Conversely, for the logarithmic score $\Psi_f=\sum_aq(a)=1$. \hfill$\blacksquare$

\subsection{Proof of Proposition~\ref{prop:within}}
\label{app:within}

By the score-function identity,
\begin{equation*}
	\begin{aligned}
	\nabla_\theta q_\theta(a\mid x)&=\sum_y\one[a(y)=a]\,\pi_\theta(y\mid x)\nabla_\theta\log\pi_\theta(y\mid x)\\
	&=q_\theta(a\mid x)\,\E_{y\sim\pi_\theta(\cdot\mid x,\,a(y)=a)}\big[\nabla_\theta\log\pi_\theta(y\mid x)\big],
	\end{aligned}
\end{equation*}
and the chain rule $\nabla_\theta L=\sum_a\frac{\partial L}{\partial q_\theta(a\mid x)}\nabla_\theta q_\theta(a\mid x)$ gives Equation~\eqref{eq:within}. \hfill$\blacksquare$

\subsection{Proof of Theorem~\ref{thm:unified}}
\label{app:unified}

Differentiating $S_\phi(q,a^*)=\phi(q)+\sum_u\frac{\partial\phi}{\partial q_u}(e_{a^*,u}-q_u)$ with respect to $q_w$,
\begin{equation*}
	\frac{\partial S_\phi}{\partial q_w}=\frac{\partial\phi}{\partial q_w}+\sum_u\frac{\partial^2\phi}{\partial q_w\partial q_u}(e_{a^*,u}-q_u)-\frac{\partial\phi}{\partial q_w}=\big[\nabla^2\phi(q)(e_{a^*}-q)\big]_w,
\end{equation*}
so $\frac{\partial}{\partial q_w}(\alpha S_\phi+\beta\Psi)=\alpha[\nabla^2\phi(q)(e_{a^*}-q)]_w+\beta[\nabla\Psi(q)]_w$. The chain rule and the score-function identity $\sum_w\frac{\partial J_x}{\partial q_w}\nabla_\theta q_w=\E_{y\sim\pi_\theta}\big[\frac{\partial J_x}{\partial q_{a(y)}}\nabla_\theta\log\pi_\theta(y\mid x)\big]$ give Equation~\eqref{eq:unified_grad}. \hfill$\blacksquare$

\subsection{Proof of Theorem~\ref{thm:maxrl}}
\label{app:maxrl}

Let $m=G-1$. By~\cite[Theorem~5]{tajwar2026maxrl}, the practical MaxRL estimator optimizes $F_m(q)=\sum_ab(a)\big(-\sum_{k=1}^m(1-q(a))^k/k\big)$ without bias, and
\begin{equation*}
	\frac{\partial F_m}{\partial q(a)}=b(a)\sum_{k=1}^m(1-q(a))^{k-1}=b(a)\,h_m(q(a)),\qquad h_m(u)\triangleq\frac{1-(1-u)^m}{u}.
\end{equation*}
At $q=b$ this is $1-(1-b(a))^m$, which is Equation~\eqref{eq:maxrl}; it strictly increases with $b(a)$, so $q=b$ violates the optimality condition unless all answers with positive probability are equally likely. $F_m$ is a sum of concave functions, so its maximizer over the simplex is characterized by the KKT conditions: there are $\lambda^*$ and $\mu_a\ge0$ with $b(a)h_m(q^*(a))+\mu_a=\lambda^*$ and $\mu_aq^*(a)=0$ for all $a$. The function $h_m$ is strictly decreasing on $[0,1]$ with maximum $h_m(0^+)=m=G-1$.
(i) \emph{Rare answers get zero probability.} If $b(a)(G-1)\le\lambda^*$ and $q^*(a)>0$, then $\mu_a=0$ and $b(a)h_m(q^*(a))<b(a)(G-1)\le\lambda^*$, a contradiction; hence $q^*(a)=0$.
(ii) \emph{Range of the multiplier.} Multiplying the KKT equality by $q^*(a)$ and summing over $a$ gives $\lambda^*=\sum_ab(a)(1-(1-q^*(a))^m)\in(0,1]$, so the threshold $\lambda^*/(G-1)$ is at most $1/(G-1)$.
(iii) \emph{Bias toward frequent answers.} For $q^*(a)>0$, $q^*(a)/b(a)=(1-(1-q^*(a))^m)/\lambda^*$. Since $h_m$ is decreasing, $q^*(a)$ increases with $b(a)$, hence so does $q^*(a)/b(a)$: frequent answers receive relatively more probability, and because $\sum_aq^*(a)=\sum_ab(a)=1$, the mode satisfies $q^*(a_0)\ge b(a_0)$. \hfill$\blacksquare$

\section{Additional diagnostic data}
\label{app:diagnostics}

\subsection{Temperature and long chain-of-thought models}
\label{app:diag}

\emph{Why $T_1$ is not greedy.} On m300, greedy accuracy is below the $T=0.6$ accuracy: 0.617 vs.\ 0.733 for \texttt{DeepSeek-R1-Distill-Qwen-1.5B} and 0.127 vs.\ 0.315 for \texttt{Qwen2.5-Math-1.5B}, because reasoning models under greedy decoding easily loop or hit the length limit.

\emph{\texttt{Qwen2.5-Math-1.5B}.} GRPO raises MATH-500 greedy accuracy from 0.094 to 0.634 in 25 steps. A method that updates only about 40 forking tokens per step also reaches 0.636 in 25 steps; 38\% of its wrong-to-right changes occur in the first 10\% of the response, mostly by changing the first tokens so that the model stops repeating the problem.

\emph{\texttt{DeepSeek-R1-Distill-Qwen-1.5B}} (60 steps, 2 seeds). Under the raw evaluation GRPO gains 1.0--5.5 points; after forcing an answer at the length limit, GRPO is only $-0.9$ to $+1.5$ points from the base model. We also designed a method aimed at responses that run past the length limit without answering. Under raw greedy evaluation it is 5.2--7.0 points above GRPO (with an 8192-token budget the fraction of greedy responses cut at the limit falls from 25.4\% to 16.9\%, at about 1/6 of GRPO's KL), but after forcing an answer it is $-0.5$ and $+0.2$ points from GRPO on the two seeds ($T=0.6$).

\subsection{Process-level diagnostics}
\label{app:process}

\emph{Failure causes.} Non-thinking Qwen models fail mostly through wrong reasoning: among wrong responses, only 0--0.4\% had earlier written a correct \verb|\boxed{}| answer (1.5--2.2\% for Qwen3). \texttt{Llama-3.2-3B-Instruct} ends 38\% of its $T=1$ responses normally without an answer box, about half of its failures. 92\% of the greedy failures of \texttt{DeepSeek-R1-Distill-Qwen-1.5B} hit the length limit, and 54.8\% of these become correct if the model is forced to answer at some segment boundary. 63\% of the greedy responses of \texttt{Qwen2.5-Math-1.5B} fall into repetition loops. \texttt{Mistral-7B-Instruct-v0.3} gets 208 of the 300 m300 problems wrong in all 16 samples.

\emph{Key steps and token-level credit.} Estimating continuation success at 8 positions of each response, the two largest changes account for 0.71--0.84 of the total change; against a smooth baseline with no key steps and only binomial noise, 24\%--56\% of correct and 51\%--68\% of wrong responses exceed the baseline's 90th percentile (10\% would be expected without key steps). A single token that turns a greedy continuation from wrong to right raises the success of $T=1.0$ continuations by only $-0.01$ to $+0.04$ relative to the original token ($+0.05$ to $+0.14$ at $T=0.6$). The variance of continuation success across segments is 1.51--3.93 times that across tokens, and within key segments only 13\%--30\% of the positions change the subsequent success by more than 0.5 when a single token is replaced. In training, the forking-token method reaches 0.590 on MATH-500 versus 0.641 for GRPO, and a method that updates only a key 32-token window lowers the KL to 1/2--1/5 of GRPO's but has the same accuracy as GRPO at equal KL.

\emph{Signals before the fact.} Token entropy identifies key steps with AUROC 0.53--0.58. Cheap text signals (disagreement of short continuations, spread of segment likelihoods, jumps of step-wise self-evaluation) reach AUROC 0.51--0.66, below the practical threshold of 0.75; on R1 the best reaches 0.61, below guessing by position (0.66), and the Spearman correlation between stated confidence and true continuation success is only 0.02--0.18. Under the same key prefix, good segments (success $\ge0.6$) and bad ones ($\le0.2$) have nearly indistinguishable mean log-probabilities (pairwise AUROC 0.45--0.51 for four non-thinking models). Linear probes on multi-layer hidden states (which read out the generated length with Spearman 0.97--0.99) identify key steps before the fact with AUROC 0.58--0.66 and pick good segments under a key prefix with AUROC 0.50--0.64; some models separate them after the fact, using the content after the step (Qwen3 0.79, Llama 0.70), but such signals cannot guide generation.

\emph{Step-wise search guided by the reference answer.} Each step samples $K$ candidate segments, estimates the success of each with $M$ verified complete continuations, and keeps the best. With $(K,M)=(2,2)$, at 0.7--2.1 times the token cost of 16 independent samples, the gap to maj@16 is $-9.6$ to $+2.8$ points; the 95\% intervals of the three Qwen models include 0, while Llama ($-7.6$) and R1 ($-9.6$) fall clearly behind. With $(K,M)=(4,4)$, at 2.8--8.3 times the cost, the three Qwen models solve 65\%--80\% of the minority problems and exceed maj@16 by 8.0--15.2 points, but all five models remain below pass@16 at 1$\times$ compute (by 1.8--4.3 points for the Qwen models, 3.5 for R1 and 15.9 for Llama).

\end{document}